\documentclass[letterpaper, 10 pt, journal, twoside]{IEEEtran}

\usepackage{amsmath,amssymb,amsfonts,url}
\usepackage{bm}
\usepackage{graphicx}
\usepackage{fontawesome5}
\usepackage{hyperref}
\usepackage{multirow}
\usepackage{booktabs}
\usepackage{color}
\usepackage{fancyhdr}
\usepackage{ifthen}
\usepackage{xcolor}
\usepackage{tabularx}
\usepackage{longtable}
\usepackage{threeparttable}

\ifCLASSINFOpdf
\else
\fi

\definecolor{bshade}{rgb}{0.55,0.75,0.95}
\newcommand{\best}[1]{%
  \colorbox{bshade!100}{\makebox(12,3)[b]{\raisebox{-0.055pt}[0pt][0pt]{\textbf{#1}}}}%
}
\newcommand{\secondbest}[1]{%
  \colorbox{bshade!40}{\makebox(12,3)[b]{\raisebox{-0.055pt}[0pt][0pt]{#1}}}%
}

\begin{document}
%
\title{Flow-Aided Flight Through Dynamic Clutters From Point To Motion}
%
%
%

\author{Bowen Xu$^{\dagger}$, Zexuan Yan$^{\dagger}$, Minghao Lu, Xiyu Fan, Yi Luo,  Youshen Lin, \\Zhiqiang Chen, Yeke Chen, Qiyuan Qiao and Peng Lu$^{*}$%
\\[0.5em]
{\small \fontfamily{ppl}\selectfont $\dagger$ Equal Contributions, * Corresponding Author}%
\\
{\small \fontfamily{ppl}\selectfont Adaptive Robotic Controls Lab (ArcLab), Department of Mechanical Engineering, The University of Hong Kong}%
\\
{\small \fontfamily{ppl}\selectfont \faEnvelope[regular] link.bowenxu@connect.hku.hk, ryan2002@connect.hku.hk, lupeng@hku.hk} \quad
{\small \fontfamily{ppl}\selectfont \color[RGB]{25,25,112} \faGithub~\href{https://github.com/arclab-hku/P2M}{arclab-hku/P2M}} \quad
{\small \fontfamily{ppl}\selectfont \color[RGB]{25,25,112} \raisebox{-0.2ex}{\includegraphics[height=1em]{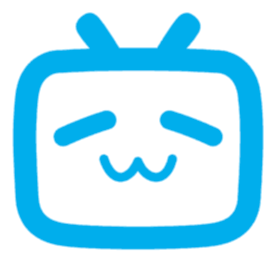}}~\href{https://www.bilibili.com/video/BV1QXm4BmEVW/}{Demo}}%
}
\maketitle

\begin{abstract}
Challenges in traversing dynamic clutters lie mainly in the efficient perception of the environmental dynamics and the generation of evasive behaviors considering obstacle movement. Previous solutions have made progress in explicitly modeling the dynamic obstacle motion for avoidance, but this key dependency of decision-making is time-consuming and unreliable in highly dynamic scenarios with occlusions. On the contrary, without introducing object detection, tracking, and prediction, we empower the reinforcement learning (RL) with single LiDAR sensing to realize an autonomous flight system directly from point to motion. For exteroception, a depth sensing \textit{distance map} achieving fixed-shape, low-resolution, and detail-safe is encoded from raw point clouds, and an environment change sensing \textit{point flow} is adopted as motion features extracted from multi-frame observations. These two are integrated into a lightweight and easy-to-learn representation of complex dynamic environments. For action generation, the behavior of avoiding dynamic threats in advance is implicitly driven by the proposed change-aware sensing representation, where the policy optimization is indicated by the relative motion modulated distance field. With the deployment-friendly sensing simulation and dynamics model-free acceleration control, the proposed system shows a superior success rate and adaptability to alternatives, and the policy derived from the simulator can drive a real-world quadrotor with safe maneuvers.
\end{abstract}

\begin{IEEEkeywords}
Aerial Systems: Perception and Autonomy; Reinforcement Learning; Autonomous Vehicle Navigation
\end{IEEEkeywords}

%
\IEEEpeerreviewmaketitle

\section{Introduction}
The past few decades have witnessed a rapid development of autonomous navigation, where elaborate mapping procedures and efficient trajectory representations \cite{wang2022geometrically} provide robust guidance for flight in complex environments and can be extended to swarm missions \cite{zhou2022swarm}. But their dependence on the perception level is still limited by compound errors introduced from inaccurate sensor observations and processing delays due to explicit environment representations. Learning-based methods attempt to alleviate these nuisances by utilizing data-driven approaches to draw complex decision-making knowledge into lightweight neural networks for high real-time performance flight through offline learning \cite{loquercio2021learning} \cite{Song2023}. Recent advances have revealed the potential for generating navigation behavior from perception to action, such as using 3D LiDAR point clouds \cite{xu2025flying}, depth image \cite{lu2024you}, and monocular optical flow \cite{hu2025seeing}, which can be transferred to the real world at a low cost.

\begin{figure}[t] \centering
	\includegraphics[width=0.485\textwidth]{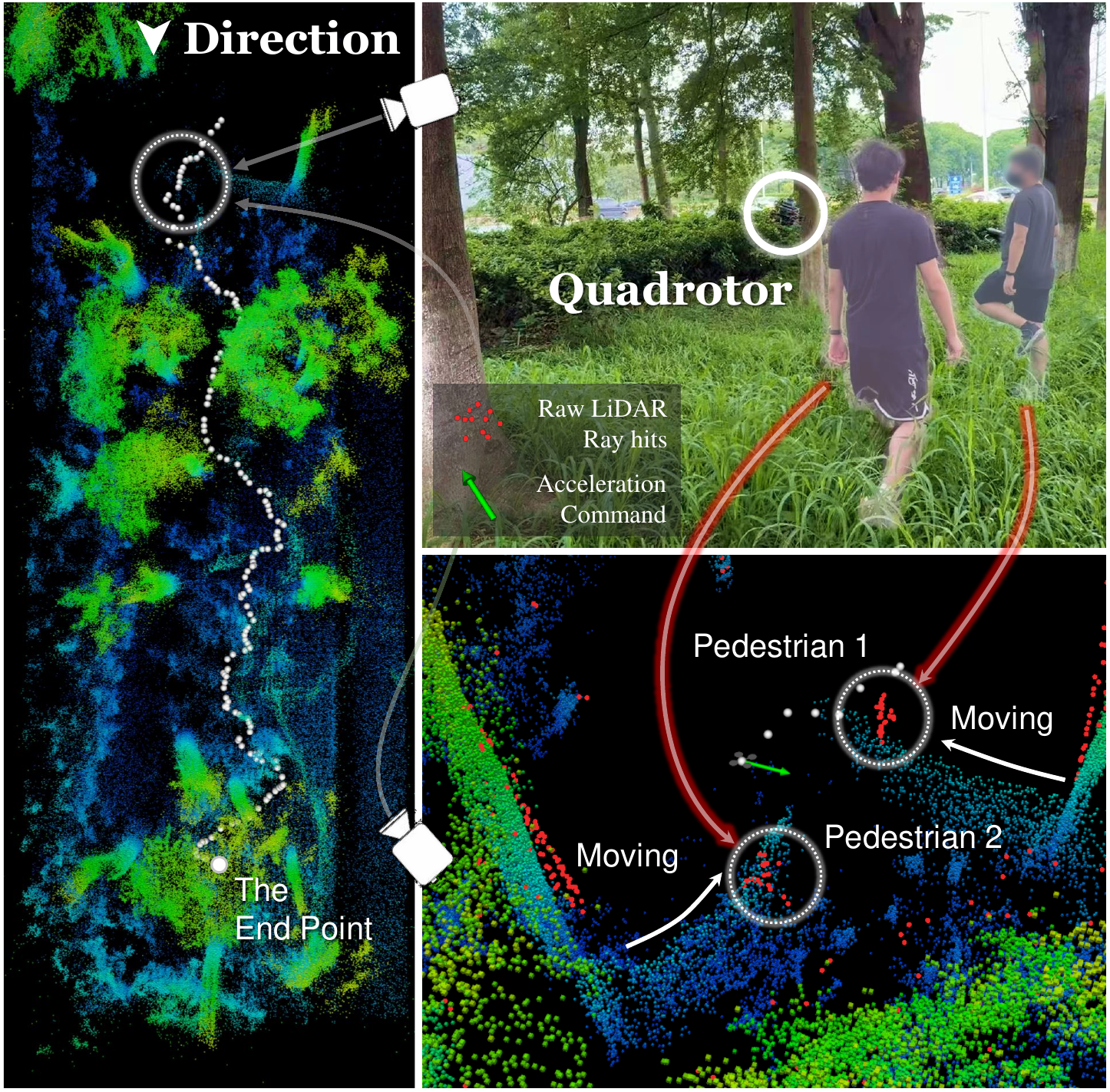}
    \caption{Outdoor flight in natural clutters with pedestrians. The left part shows the executed trajectory of the neural controller. The snapshots in the right part demonstrate the avoidance response when confronted with both dynamic pedestrians and static trees. The movement of the pedestrians can be seen clearly in the “ghost trail” of the map built after the experiment.}
	\label{fig_cover}
\end{figure}

The presence of dynamic objects makes it difficult for the robot to determine the complete environment occupancy by compensating for ego-motion. In order to intuitively account for obstacle movement in the planning session, existing works choose to distinguish dynamic objects from the environment. However, occluded motion in onboard observations constitutes an important component of sensed environmental dynamics, but object-level understanding \cite{lu2022perception} requires consistent and coherent localized characteristics and motion models. This causes potential neglect of incomplete motions. Furthermore, the simultaneous state estimation of multiple objects \cite{lu2024fapp} creates a huge computational burden, and sensor fusion strategies \cite{xie2023drl} exacerbate the perceptual latency. More importantly, the hand-crafted detection-tracking process requires multiple processes to work in concert, introducing inter-module compound errors \cite{xu2025navrl}. The above factors make the dynamic perception of object-aware approaches untimely and unreliable as inputs to subsequent modules, e.g., state predictions for trajectory optimization or network policies for action generation.

In this letter, we present an autonomous flight system that leverages LiDAR sensing in conjunction with a generalizable RL technique. Considering that using raw points for training is computationally inefficient, and oversampling leads to a loss of obstacle information, we design a low-resolution distance map with fixed shape and preserving detail safety based on the robot-centric nature of LiDAR scans. For dynamic environments, we aim to find a fine-grained representation that is correlated with environmental changes, and is consistent with the distance map in scale, shape, and positional meaning, which naturally guides us to supplement pixel motion for each distance sensing. We use this point-based relative motion feature \textbf{point flow} obtained from a pre-trained neural optical flow estimator, together with the depth perception \textbf{distance map} as the policy input. We implement the training by encouraging evasive maneuvers against dynamic obstacles that are implicitly driven by the change-aware input, which is correlated to the obstacle relative motion during policy optimization. The stable LiDAR-inertial proprioception, compact and realistically portable sensing simulation, and end-to-end policy architecture allow the system to be applied to physical aerial robots. 

Our main contributions can be summarized as follows:
\begin{itemize}
    \item \textit{Propose an efficient LiDAR representation combining the depth and environment change sensing}, where a distance map achieves low-resolution depth perception while preserving detail safety, and a point flow describes the holistic relative motion of the surroundings.
    \item \textit{Develop a generalizable RL training for dynamic obstacle avoidance} through learning early avoidance maneuvers from implicit environmental motion features, where the policy optimization is hinted by the relative motion-reshaped distance field of the dynamic obstacle.
    \item \textit{Integrate a lightweight autonomous flight system from point to motion} without the need of object detection, tracking, and prediction. Extensive simulations and real-world experiments verify its higher success rate than baselines, especially in complex dynamic environments.
\end{itemize}
\section{Related Work}
\label{sec:related_work}
\subsection{Hierarchical Object-aware Dynamic Obstacle Avoidance}
Technical routes to discriminate dynamic objects from the environment for obstacle avoidance have been widely used. Some reactive solutions are based on repulsive potential field from moving object events \cite{falanga2020dynamic} and trajectory from model predictive control (MPC) \cite{xu2022dpmpc}, but the in-the-moment optimization only focuses on the local optimum of safety and ignores the long-term requirements of navigation tasks such as smooth movement and heading to the goal, resulting in low adaptability and efficiency in complex dynamic environments. More complete systems use both dynamic clusters and static occupancy for trajectory optimization, where rgb-d based system \cite{lu2022perception} projects small object recognition results into the depth domain and point cloud based systems \cite{wang2021autonomous} \cite{lu2024fapp} cluster and segment the spatial information, these methods estimate the states of dynamic obstacles and further predict their state sequences for calculating the cost function of planning. However, the comprehensive modeling of dynamic objects introduces compound errors for the final prediction module, which leaves the trajectory optimization module to rely on inaccurate environmental references. Building on these efforts, learning from object recognition is adopted to bypass the unreliable prediction-planning pipeline, where \cite{xie2023drl} uses the encoded detection result image as pedestrian information and \cite{xu2025navrl} utilizes obstacle state vectors as dynamism sensing. But fundamentally limited by the object-aware strategy, their detection-tracking modules underutilize the environment dynamics for struggling to account for the incomplete motions caused by occlusions, and their real-time performance decreases drastically when the quantity of dynamic obstacles increases, thus failing to maintain safety in dynamic and cluttered scenarios. In comparison, our method does not scrutinize the dynamic-static distinction in sensing, where the environmental changes throughout the observation can be uniformly represented at an object-free pixel level for end-to-end evasive maneuvers optimization.

\subsection{End-to-end Navigation in Dynamic Clutters}
For dynamic environment navigation that maps perception to action, existing approaches leverage the high-frequency inference properties of neural controllers for adaptability to changing environments, where \cite{zhang2025learning} learns with differentiable simulations using low-resolution depth images, and \cite{perez2021robot} derives velocity and steering angle from single-frame LiDAR scans. These works rely on static sensing that achieves avoidance of simple moving objects, but is difficult to generalize to complex dynamic environments. This is because the lack of historical information prevents them from sensing additional threats introduced by obstacle motion. To enable dynamic awareness, learning from past sensing is presented. Velocity command of \cite{fan2020distributed} is mapped from the last three frames of LiDAR scans, and acceleration control of \cite{fan2025flying} is achieved with an obstacle image stitched from past scan vectors. However, motion patterns in multi-frame observations are difficult to capture without explicit supervision. This lack of emphasis on inter-observation temporal variations makes it difficult to learn safety maneuvers triggered by obstacle movement patterns, which can lead to overfitting of the training dynamic scenes. On the basis of the above works, integrating multi-frame perception is proposed to obtain more explicit motion-related inputs, where \cite{de2024spatiotemporal} represents obstacle kinematics from point cloud alignment results and \cite{hoeller2021learning} denotes scene temporal evolutions as long short-term memory-encoded depth images. For learning avoidance behaviors, these representations are of good benefit, but obtaining them requires specially optimized feature extraction modules. Training these modules from scratch entails expensive customized datasets, whereas co-optimization in RL leads to unstable and unnecessary updating of the dynamic perception module subject to policy gradients, making it difficult to learn generic features. In contrast, our approach utilizes pre-trained weights to align the geometric structure of historical observations with the semantics of pixel motion, obtaining a low-noise representation which is highly correlated with the subtask space of environmental changes. The flow attention serves as an aid to depth perception, providing a reliable environment representation which is further embedded to guide navigation decisions.
\section{Framework}

\subsection{Distance Map Encoding}
\label{Distance Map Encoding}
Given the LiDAR horizontal and vertical field of view (FOV) and their boundaries as $F_H  \in (F_{Hh}, F_{Hl}),\ F_V  \in (F_{Vh}, F_{Vl})$, we consider all point clouds within the truncation distance $d_{\text{max}}$. We define a current scanning point set in the body frame as $_{B}P$, and for each point $p_i \in {}_{B}P$, it has a coordinate $(x_i, y_i, z_i) $ and a distance to the quadrotor $d_i = \sqrt{x_i^2 + y_i^2 + z_i^2}$. A raycast operation is first applied to organize the raw point cloud into an ordered point set with a fixed shape of $r_h \times r_v \times 3 = 108 \times18\times 3$, where space within the FOV is divided by intervals of $i_h = F_H/r_h, i_v = F_V/r_v$ in the horizontal and vertical direction. Each point can determine which partition it belongs to according to its azimuth $\theta_h$ and elevation $\theta_v$ relative to the quadrotor. The point within the $s$th ($m$th in horizontal, $n$th in vertical, $s = r_h(n-1)+m$) partition is $P_s = \{p_i \in  {}_{B}P \ |\ \xi\}$.
\begin{equation}
\begin{aligned}
&\quad \theta_{hi} = \tan^{-1}(y_i/x_i), \ \theta_{v_i} = \sin^{-1}(z_i/d_i),\\
\xi := & \frac{\theta_{hi}-F_{Hl}}{i_h} \in (m-1, m), \frac{\theta_{vi} - F_{Vl}}{i_v} \in (n-1, n).
\end{aligned}
\label{eq:lidar_direction}
\end{equation}

For the points in each partition, we take the point with the smallest distance value as the representative point of this partition. This approach retains more obstacle details than simply downsampling, essentially inflating the nearest obstacle point in its partition to improve safety. 
The point $p_s$ corresponding to the $s$th partition in the raycast point set can be selected as $p_s = \arg \min_{p_i \in P_s}d_i$. 

Then, to obtain the image-form depth representation, we first remove unnecessary detections from the ceiling or ground, only keeping the point cloud within a relative height interval of the quadrotor. Then, we take the distance value of each partition of the raycast point set to encode a high-resolution ($108 \times18$) grayscale image. A further minimal pooling is adopted to get the low-resolution ($36 \times6$) distance map.

\begin{figure} \centering
	\includegraphics[width=0.485\textwidth]{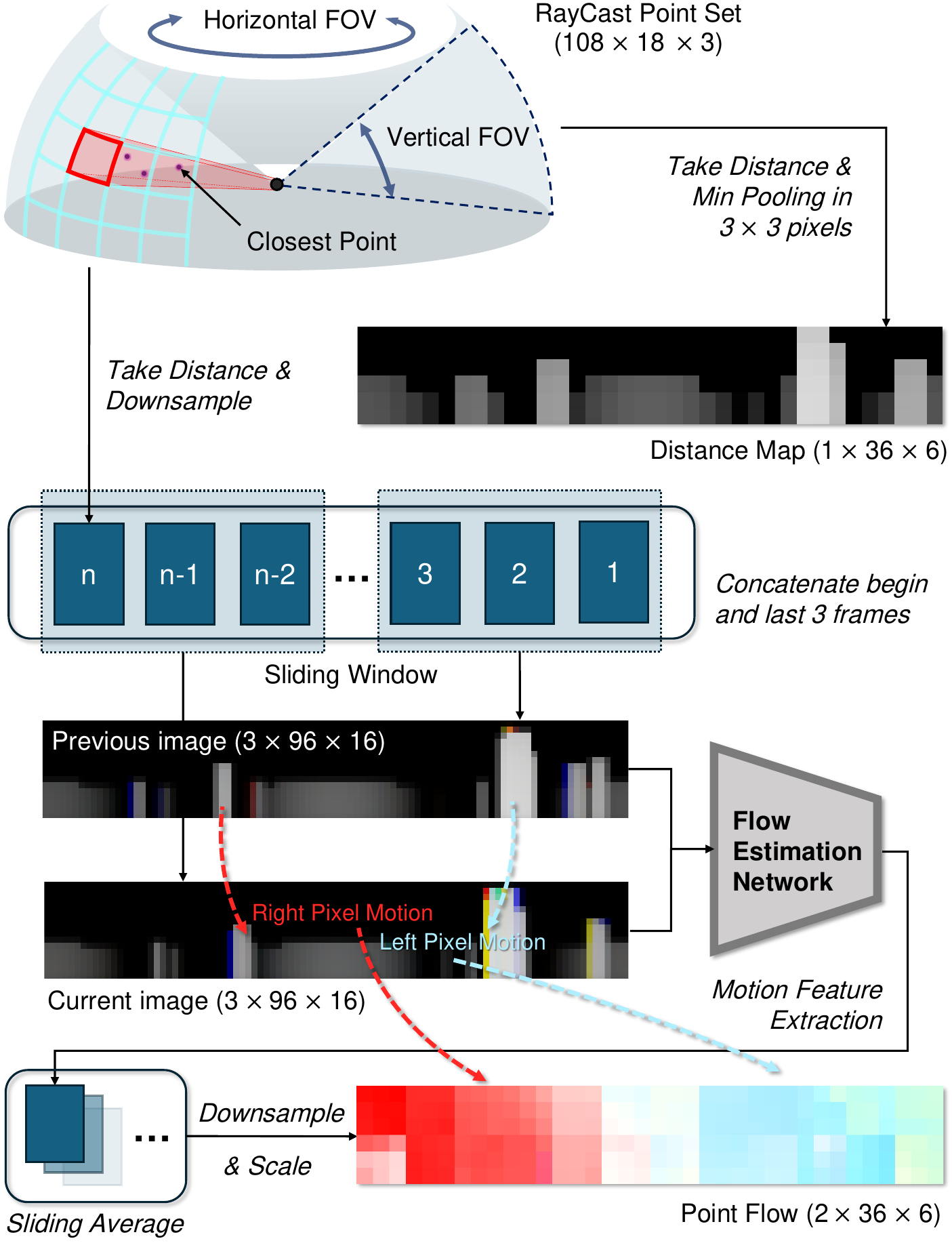}
    \caption{Illustration of the change-aware LiDAR sensing representation. The distance map is encoded by robot-centric omnidirectional depth perception, which ensures safe sensing of obstacle details through nearest selection. The point flow is derived from pseudo-RGB image pairs stacked with 3 consecutive grayscale images, which are fed into a flow estimation module that does not update during RL training for motion feature extraction.}
	\label{fig_distance_map_and_point_flow}
\end{figure}

\subsection{Point Flow Estimation}
To enable parallel flow computation in training, and inference acceleration in testing, we use a learning-based method, NeuFlowV2 \cite{zhang2024neuflow}, to provide real-time and robust flow estimation for our system.
As shown in Fig. \ref{fig_distance_map_and_point_flow}, we resize the grayscale image encoded from the raycast point set from ($108 \times18$) to ($96\times16$) using bilinear interpolation, and combine three consecutive frames into a single 3-channel image to fit the input of the pre-trained Neuflow checkpoint. Then, a sliding window with length $n$ is applied to ensure continuous and high-frequency input. We consider the previous concatenated image $I_{k-n+3}$, which is $n-3$ frames ahead of the current concatenated image $I_{k}$. The previous frame pixels $(u_{k-n+3}, v_{k-n+3})$ are re-projected to the current frame pixels $(u_k, v_k)$ after relative motion with the quadrotor. The real flow $F_k$ can be estimated as $\hat{F_k}=\text{neuflow}(I_{k-n+3}, I_k)$ and a sliding average is carried out on 5 frames of continuous flow with a window size of $5/50\ \text{Hz}=0.1\ \text{s}$ to balance the estimation error. 

After downsampling the average flow $\bar{F}_{k \rightarrow k-4}$ to the low-resolution form $\bar{F}'_{k \rightarrow k-4}$ to match the size of the distance map, we can obtain the point flow by further scaling:
\begin{equation}
\begin{aligned}
 &\quad \bar{F}'_{k \rightarrow k-4} = (\bar{\Delta u}'_{k \rightarrow k-4}, \bar{\Delta v}'_{k \rightarrow k-4})_{2 \times 36 \times 6},  \\
 &\text{pointflow}_{k} = (\frac{\bar{\Delta u}'_{k \rightarrow k-4}}{z_H}, \frac{\bar{\Delta v}'_{k \rightarrow k-4}}{z_W})_{2 \times 36 \times 6},\\
\end{aligned}
\label{eq:optical_flow_estimation}
\end{equation}
where the horizontal and vertical flow values of $\bar{F}'_{k \rightarrow k-4}$ are scaled by $z_H$ and $z_W$ respectively to re-balance the attention on the pixel motion of different directions.

\subsection{System Overview of Observation and Action}
\begin{figure*} \centering
	\includegraphics[width=0.87\textwidth]{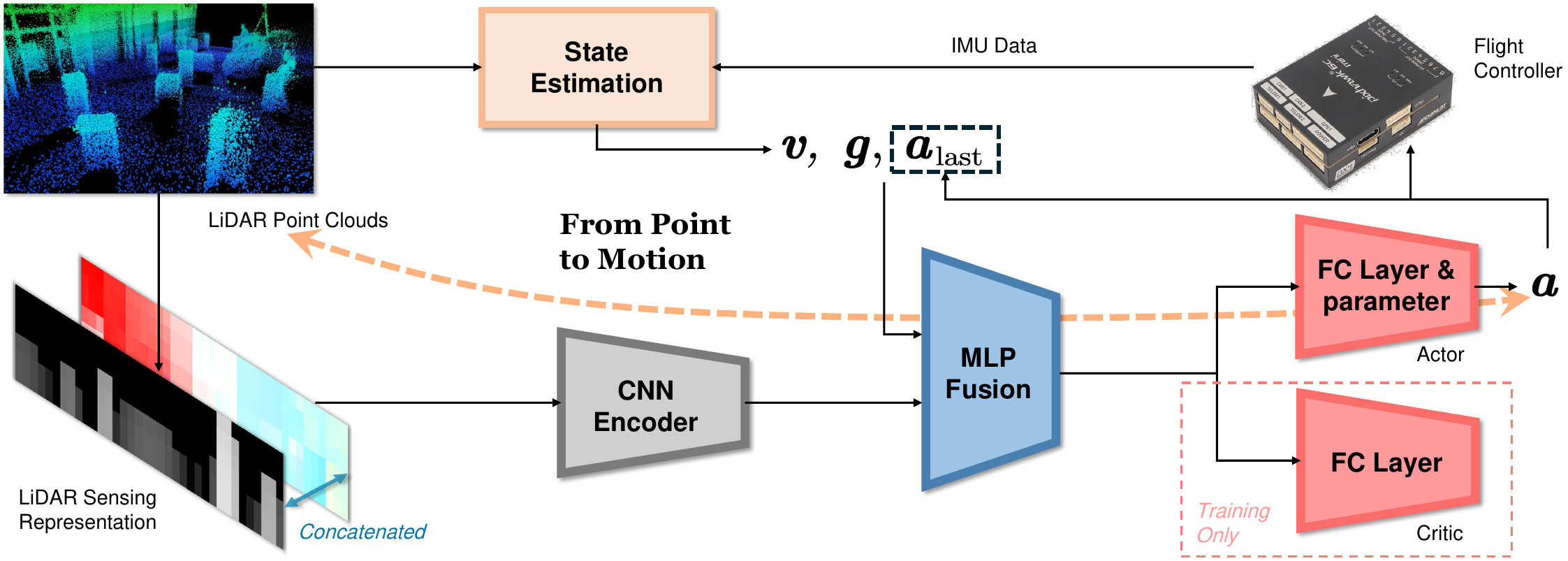}
	\caption{Implementation of the autonomous flight system from point to motion. State estimation \cite{xu2022fast} of the quadrotor is realized through the collaborative use of the LiDAR point clouds and the IMU data, which is a robust real-world substitute for ground truth states used in training. The unit vector pointing to the goal $\boldsymbol{g}$, the velocity $\boldsymbol{v}$ and the last output $\boldsymbol{a}_{\text{last}}$ form the state inputs of the system, which are used to feed into the MLP fusion module together with the LiDAR sensing representation features extracted by the CNN encoder. The output of the system is the desired acceleration command $\boldsymbol{a}$.}
	\label{fig_system_overview}
\end{figure*}
The framework of the system is shown in Fig. \ref{fig_system_overview}. The distance map and the point flow are concatenated in the channel dimension into a tensor of shape (3,36,6) for working together as the LiDAR sensing representation. The unit vector pointing to the goal $\boldsymbol{g}$, the velocity $\boldsymbol{v}$, and the last action $\boldsymbol{a}_{\text{last}}$ are all 3D vectors with the shape of (3,), which inform the target direction and ego-motion. A network policy is used for acceleration command inference of (3,). These system designs select both observation and action to be at the kinematic level, thereby achieving a 
universal navigation method enabling policy deployment on different quadrotors across training and reality, regardless of the inconsistent underlying dynamics parameters.

\section{Reinforcement Learning for the Obstacle Avoidance Policy}

\subsection{Policy Representation}
We employ a partially shared architecture to model the policy and the value functions, which share an encoder based on convolutional neural network (CNN) and fully connected (FC) layer. The convolutional part has three layers, whose structure [output channel, kernel size, stride, padding] is respectively [4, (5,3), (2,1), 0], [16, (5,3), (2,1), (2,1)]$\times2$. The output of the CNN layer is flattened and is then fed into the FC layer of 128 units and normalized for extracting LiDAR features. After being concatenated with state inputs, the observations are fused by the multi-layer perception (MLP) of [256, 256] and fed into an actor-critic module.

\subsection{Reward Function}
\label{reward_function}
The reward function is designed according to the safety-assured, goal-oriented, and dynamically feasible requirements of navigation tasks, and a dynamic obstacle avoidance reward is adopted for better adaptation to the dynamic environment. The reward function $r$ can be described as:
\begin{equation}
r = r_{\text{state}} + r_{\text{goal}} + r_{\text{safety}} + r_{\text{dobs}},
\label{eq:total_reward}
\end{equation}
where $r_{\text{state}}$, $r_{\text{goal}}$, $r_{\text{safety}}$ and $r_{\text{dobs}}$ represent the state reward, goal reward, safety reward and dynamic obstacle avoidance reward.

\emph{1) State reward}: 
The quadrotor states are softly constrained to the desired range by a limiting function:
\begin{equation}
\begin{aligned}
f_{l}&(s, \beta_s, s_{\text{low}}, s_{\text{high}}) \\ 
& = \log(e^{-\beta_s(\max(s_{\text{low}} - s, 0) + \max(s-s_{\text{high}}, 0))} + 1),
\end{aligned}
\label{eq:limiting_function}
\end{equation}
where $s$, $\beta_s$, and $[s_{\text{low}},s_{\text{high}}]$ represent the quadrotor state magnitude, the corresponding discount factor, and the desired range. Then, we define a velocity reward $r_{v}$, an acceleration reward $r_{a}$, a height reward $r_{h}$ to encourage the desired flight states, and a jerk reward $r_{j}$ to promote a smooth trajectory. 
\begin{equation}
\begin{aligned}
& r_{v} = f_{l}(\|\boldsymbol{v}\|, \beta_v, v_{\text{min}}, v_{\text{ref}}), \  r_{a} = f_{l}(\|\boldsymbol{a}\|, \beta_a, a_{\text{min}}, a_{\text{ref}}), \\ 
& r_{h} = f_{l}(p_z, \beta_h, h_{\text{min}}, h_{\text{max}}), \  r_{j} = 1/(1+\|\boldsymbol{a}_t - \boldsymbol{a}_{t-1}\|), \\
& \qquad \qquad r_{\text{\text{state}}} =  \lambda_vr_{v} + \lambda_ar_{a} + \lambda_hr_{h} + \lambda_jr_{j},
\end{aligned}
\label{eq:state_reward}
\end{equation}
where $\boldsymbol{v}$ and $\boldsymbol{a}$ are the quadrotor velocity and the acceleration command. $v_{\text{ref}}$ $(a_{\text{ref}})$ and $v_{\text{min}}$ $(a_{\text{min}})$ are the reference and minimum velocity (acceleration), which define the expected and minimum-acceptable agility. $p_z$ is the $z$ coordination of the quadrotor position $\boldsymbol{p}$, $h_{\text{min}}$ and $h_{\text{max}}$ are the bounds of the appropriate flight height, and $\lambda$ is the relative weight.

\emph{2) Goal reward}: 
The behavior of flying to the goal is guided from two perspectives: the projection of the velocity in the goal direction $r_{\text{dir}}$ and the change of distance to the goal $r_{\text{dis}}$, where $r_{\text{dir}}$ encourages a goal-aligned velocity direction and $r_{\text{dis}}$ encourages an actual flight progress.
\begin{equation}
\begin{aligned}
&r_{\text{dir}} = \min(\boldsymbol{v} \cdot \boldsymbol{p}_g/\|\boldsymbol{p}_g\|, v_{tr}), \ r_{\text{dis}} = e^{d_{g}(t-1) - d_{g}(t)} - 1, \\
& \qquad \qquad \qquad r_{\text{goal}} =  \lambda_{\text{dir}}r_{\text{dir}} + \lambda_{\text{dis}}r_{\text{dis}}, 
\end{aligned}
\label{eq:goal_reward}
\end{equation}
where $\boldsymbol{p}_g$ and $d_{g}$ are the goal position and the distance to the goal. The truncated projection velocity $v_{tr} = 0.4v_{\text{ref}}$ promotes a later training stage that shifts the dominance of $r_{\text{goal}}$ from the saturated $r_{\text{dir}}$ to $r_{\text{dis}}$, towards a safety-balanced goal-directed skill than the initial velocity alignment.

\emph{3) Safety reward}: 
Safety reward is designed as a two-stage reward that can be derived directly from LiDAR scans.
\begin{equation}
\begin{aligned}
d_s =&
\begin{cases} 
\operatorname{avg}\{d_{li}-r_d\ | \ d_{li} < d_{th}\}, & \min(d_{li}) < d_{th} \\
 \min(d_{li}-r_d), & \text{otherwise}
\end{cases}  \\
&\qquad r_{\text{safety}} = \lambda_s\max(\log(d_s), \log(d_{tr})),
\end{aligned}
\label{eq:safety_reward}
\end{equation}
where $d_{li}$ and $r_d$ are the distance of the $i$th ray hit and the radius of the quadrotor, $d_{\text{th}}=1$ is the distance threshold. The truncated logarithmic distance $\log(d_{tr}) =-5$ means that when the agent is close to the obstacle, extreme negative values should be avoided to dominate the total reward.

\begin{figure} \centering
	\includegraphics[width=0.485\textwidth]{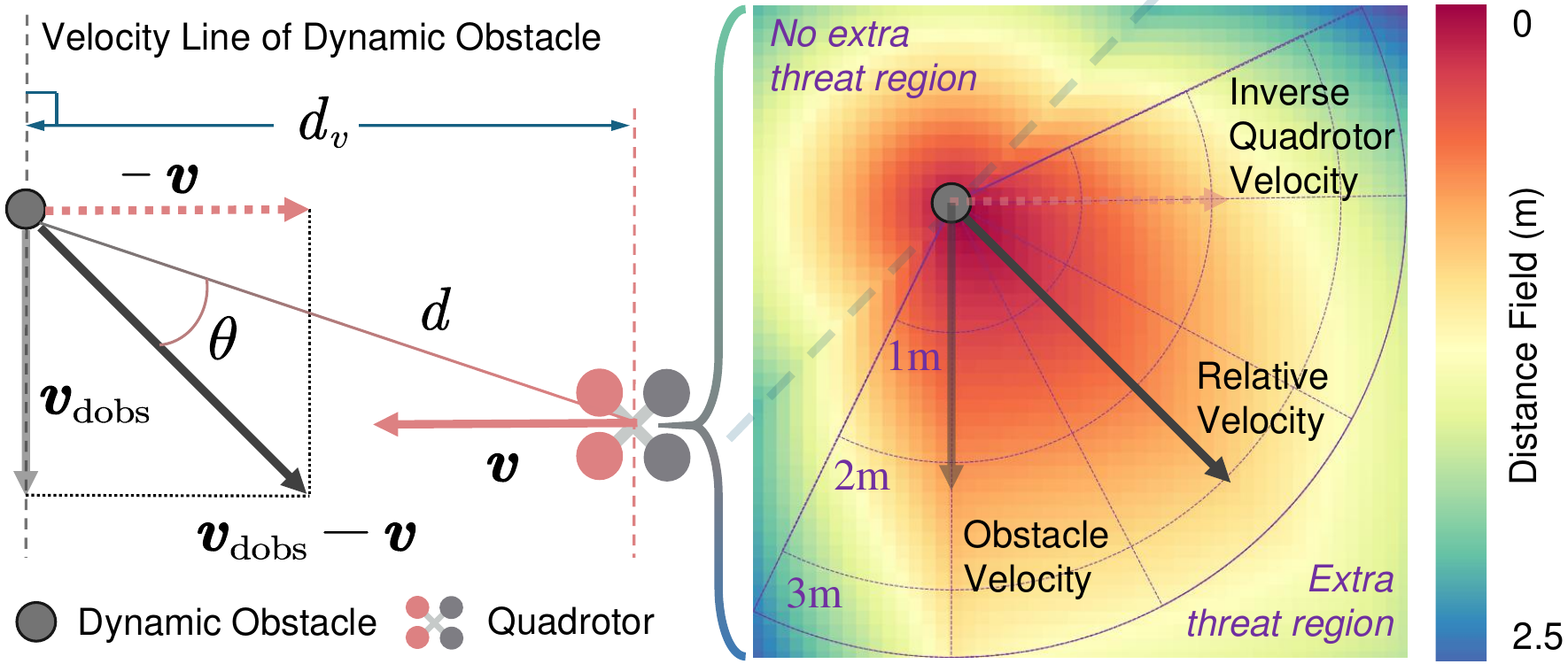}
    \caption{Effects of reshaping the distance field of dynamic obstacles. The distance field is shown in a range of $5 \times 5\ m^2$ when the dynamic obstacle is at the same orthogonal velocity of $1\ m/s$ as the quadrotor. This exhibits the extra threat given to $\theta\in (0, \pi/2)$ is roughly symmetrical about the relative velocity $\boldsymbol{v}_{\text{dobs}} - \boldsymbol{v}$ and skewed toward the absolute obstacle velocity $\boldsymbol{v}_{\text{dobs}}$.}
	\label{fig_dobs_reward}
\end{figure}

\emph{4) Dynamic obstacle avoidance reward}: 
The dynamic obstacle avoidance reward is considered on a horizontal level and realized by reshaping the distance field of the dynamic obstacles, which aims to encourage the quadrotor to avoid potential danger in advance. The extra collision threat region of a dynamic obstacle is defined by the following condition.
\begin{equation}
\theta = \cos^{-1}(\frac{(\boldsymbol{p}-\boldsymbol{p}_{\text{dobs}}) \cdot (\boldsymbol{v}_{\text{dobs}} - \boldsymbol{v})}{\|\boldsymbol{p}-\boldsymbol{p}_{\text{dobs}}\| \cdot \|\boldsymbol{v}_{\text{dobs}} - \boldsymbol{v}\|}), 
\label{eq:danger_dobs}
\end{equation}
where $\boldsymbol{p}_{\text{dobs}}$ and $\boldsymbol{v}_{\text{dobs}}$ are the position and velocity of a dynamic obstacle. Eq. \eqref{eq:danger_dobs} evaluates the angle between the relative velocity (obstacle velocity in the quadrotor body frame) and the connection between the obstacle and the quadrotor. Then the reshaping coefficient $k$ can be defined with $\theta$ as follows:
\begin{equation}
\begin{aligned}
&\ \ \ \  k_v = \| \boldsymbol{v}_{\text{dobs}} - \boldsymbol{v} \|, \ k_{\theta} = 1 - (\frac{2\theta}{\pi}), \ k_d = e^{\frac{1}{1+d_v}}, \\
&k = 1 + k_v \cdot k_{\theta} \cdot  k_d \ \ \text{if} \ \theta \in (0, \frac{\pi}{2}),\ \ \text{otherwise} \ k= 1,
\end{aligned}
\label{eq:reshape_coef}
\end{equation}
where $d_v$ represents the distance from the quadrotor to the obstacle velocity line, and $k$ is used to discount the original distance field as in Fig. \ref{fig_dobs_reward}. 

Considering only the dynamic obstacles hit by the rays is reasonable, but this will make the reward highly random due to the fast-changing nature of dynamic surroundings. One way to stabilize it is to relax the hitting constraint and replace it with considering all dynamic obstacles within a certain distance to the quadrotor. Although privileged information is introduced, this solution can help better convergence and can be easily generalized to scenarios with occlusions.
\begin{equation}
r_{\text{dobs}} =
\begin{cases} 
\dfrac{\lambda_d}{m}\sum_{m}\max(\log(\dfrac{d_{\text{dobs}i}-r_d}{k_i}), \log(d_{tr})), m > 0&  \\
 \max(r_{\text{dobs}}(t-1), \ ...,\ r_{\text{dobs}}(0)),  \quad \quad \ \ \ \ \ \text{otherwise}&
\end{cases}
\label{eq:dobs_reward}
\end{equation}
where $d_{\text{dobsi}}$ is the distance between the $i$th dynamic obstacle and the quadrotor, $m$ is the number of dynamic obstacles within $0.75d_{\text{max}}$. When no dynamic obstacles are in this range, the reward is set to the maximum of the previously occurring $r_{\text{dobs}}$ to encourage this behavior of totally away from dynamic threats, while reducing the reward sparsity.

\subsection{Training Setup and Implementation}
\begin{figure} \centering
	\includegraphics[width=0.485\textwidth]{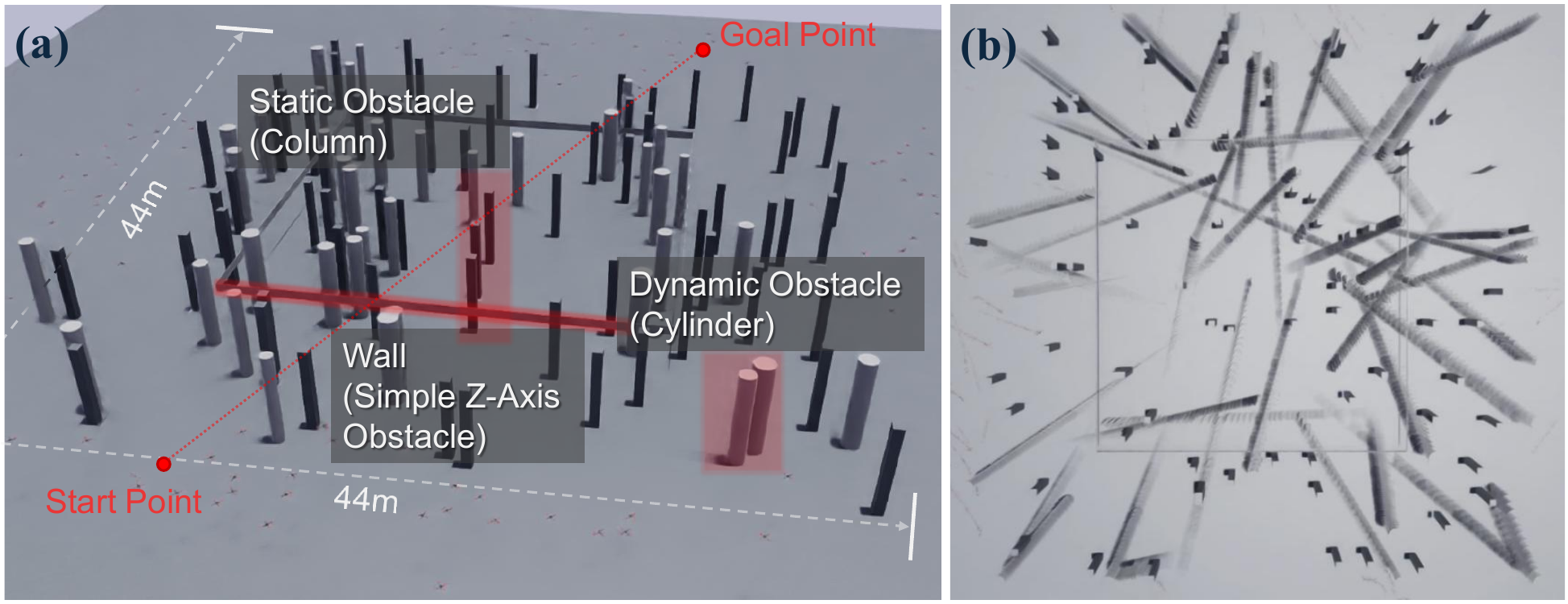}
    \caption{Training Environment and Obstacle Setup. (a) Parallel training in 128 environments with centrosymmetrically set start-goal points, and walls are set at the same height as the start-goal points. (b) Top-down view of the trajectories of the dynamic obstacles performing uniform linear motions.}
	\label{fig_training_env}
\end{figure}

As shown in Fig. \ref{fig_training_env}, static columns and walls are set up to guide 3D avoidance actions. Dynamic obstacles are set to move in uniform linear motions with a randomly assigned velocity within $[1, 5]\ m/s$. We define early termination for (i) agent exceeding the limit of velocity ($1.2v_{\text{ref}}$), acceleration ($1.5a_{\text{ref}}$) or height, (ii) being too close to obstacles, and (iii) straying too far from the flight course. The max episode length is set as higher than the process buffer required for flying to the goal, and $r_{\text{dis}}$ and $r_{v}$ are cut off when the quadrotor is close enough to the goal, changing the training mode from a navigation task to a safe wandering around the goal. The implementation of the underlying proximal policy optimization (PPO) \cite{schulman2017proximal} training toolchain is highly inspired by Omnidrones \cite{xu2024omnidrones}, which is built on Nvidia Isaac Sim.

\section{Experiments and Evaluations}

\subsection{Simulation Experiments}
\emph{1) Benchmark with previous systems}:
We benchmark our system against recent representative works: a hierarchical rule-based method FAPP \cite{lu2024fapp}, a hierarchical learning-based method NavRL \cite{xu2025navrl}, and an end-to-end method Obsnet \cite{fan2025flying}, all of which serve as the state-of-the-art navigation approaches. For a reasonable comparison, we set the reference acceleration to $10\ m/s^2$ for acceleration control-based \cite{fan2025flying} and the proposed method, and the acceleration limit of \cite{lu2024fapp} to $10\ m/s^2$. In terms of velocity, \cite{fan2025flying} does not need a reference velocity input; the proposed method accepts $v_{\text{ref}}$ of $5\ m/s$; and the local greedy trajectory optimization design of \cite{lu2024fapp} requires a more stringent velocity limit, which is set to $3.5\ m/s$. For \cite{xu2025navrl} trained towards low-speed scenarios, its velocity command output is set within $[0.5, 1.5]\ m/s$, and its perception range is also reduced to $5\ m$ compared to $10\ m$ for other methods to match its agility.

\begin{figure} \centering
    \includegraphics[width=0.485\textwidth]{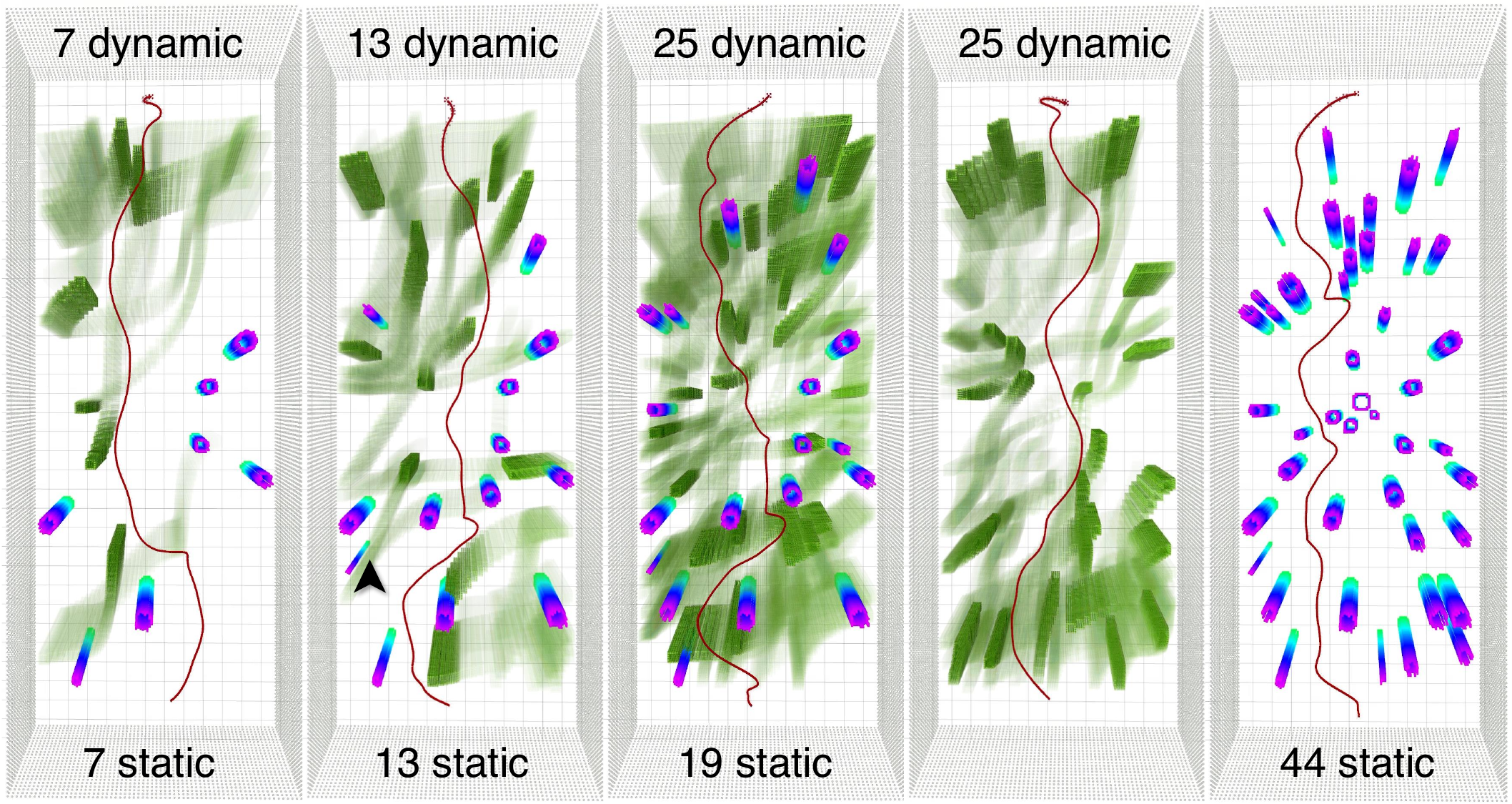}
    \caption{Simulated flight slices of the proposed system. The benchmark simulation experiments are executed in a $25 \times 10\ m^2$ field, where static obstacles contain both thick and thin columns, and the maximum speed of dynamic obstacles is limited to below $5\ m/s$.}
	\label{fig_baseline_compare}
\end{figure}

The simulation environment is shown in Fig. \ref{fig_baseline_compare}, where the dynamic obstacles perform variable-speed curve motion, which is unseen compared to that in training. We counted the success rate $\eta$ from $20$ experiments and other performance metrics from $5$ successful flights. The evaluation of performance metrics is performed on an i5-14600KF CPU and RTX4070Tis GPU, including the average flight speed $v_a$, planning latency from perception to action $t_p$, path efficiency $R_l=l/\|\boldsymbol{p}_g - \boldsymbol{p}_s\|$ meaning the path length $l$ divided by the distance from start $\boldsymbol{p}_s$ to goal $\boldsymbol{p}_g$, and safety distance $d_s$.
\begin{table}[t]
    \setlength{\tabcolsep}{5pt}
    \centering
    \caption{Baseline Comparison in Cluttered Environments with \\ Different Densities of Static and Dynamic Obstacles}
    \begin{threeparttable}   
    {
    \begin{tabular}{m{1.3cm}<{\centering}m{0.9cm}<{\centering}m{0.6cm}<{\centering}m{1.0cm}<{\centering}m{0.8cm}<{\centering}m{0.6cm}<{\centering}m{0.7cm}<{\centering}}
       \toprule
       \toprule
       \textbf{Scenario} & \textbf{Method} & \textbf{$\eta(\%)$} & \textbf{$v_a(m/s)$} & \textbf{$t_p(ms)$} & \textbf{$R_l$}& \textbf{$d_s(m)$} \\
       \hline
       
     \multirow{3.5}{*}{7 dynamic}
     & Ours & \best{95} & \secondbest{3.14} & 9.58 & 1.13 & \best{2.80}\\
     \multirow{3.5}{*}{and 7 static}
     & \cite{lu2024fapp}& \secondbest{90} & 2.20 & \secondbest{9.31} & 1.08 & \secondbest{2.43}\\ 
     & \cite{xu2025navrl}& 45 & 0.88 & 24.30 & \best{1.04} & 1.71\\
     & \cite{fan2025flying}& 20 & \best{5.73} & \best{4.05} & \secondbest{1.05} & 0.97\\
     \hline
     
     \multirow{3.5}{*}{13 dynamic}
     & Ours & \best{65} & \secondbest{2.55} & \secondbest{10.38} & 1.43 & \best{2.05}\\
     \multirow{3.5}{*}{and 13 static}
     & \cite{lu2024fapp}& \secondbest{30} & 1.88 & 12.75 & 1.22 & \secondbest{1.93}\\
     & \cite{xu2025navrl}& \secondbest{30} & 0.89 & 26.19 & \best{1.06} & 1.62\\
     & \cite{fan2025flying}& \secondbest{30} & \best{5.61} & \best{5.06}  & \secondbest{1.15} & 0.90 \\ 
     \hline 

     \multirow{3.5}{*}{25 dynamic}
     & Ours & \best{40} & \secondbest{2.29} & \secondbest{12.29} & \secondbest{1.20} & \best{1.83}\\
     \multirow{3.5}{*}{and 19 static}
     & \cite{lu2024fapp}& \secondbest{25} & 2.09 & 21.36 & 1.21 & \secondbest{1.74}\\
     & \cite{xu2025navrl}& 20 & 0.90 & 33.85 & \secondbest{1.20} & 1.39\\
     & \cite{fan2025flying}& 20 & \best{5.35} & \best{6.92}  & \best{1.08} & 0.93\\ 
     \hline 

     \multirow{4.5}{*}{25 dynamic}
     & Ours & \best{50} & \secondbest{2.92} & \secondbest{10.26} & 1.27 & \best{2.30}\\
     & \cite{lu2024fapp}& 20 & 1.94 & 17.01 & 1.31 & \secondbest{2.05}\\     
     & \cite{xu2025navrl}& 15 & 0.87 & 28.28 & \secondbest{1.25} & 1.48\\
     & \cite{fan2025flying}& \secondbest{25} & \best{5.33} & \best{4.32} & \best{1.07} & 0.96\\
       \hline 
     
     \multirow{4.5}{*}{44 static}
     & Ours & 60 & \best{2.15} & \best{8.88} & 1.24& \best{1.52}\\
     & \cite{lu2024fapp}& \secondbest{75} & \secondbest{1.84} & \secondbest{22.89} & \best{1.07} & \secondbest{1.36} \\ 
     & \cite{xu2025navrl}& \best{80} & 0.92 & 27.09 & \secondbest{1.13} & 1.32 \\
     & \cite{fan2025flying}& / & / & / & / & /\\
     \bottomrule
     \bottomrule
    \end{tabular}
    }
    \end{threeparttable}
    \label{tbl:baseline_compare}
\end{table}

The perception modules in \cite{lu2024fapp} and \cite{xu2025navrl} work well in low obstacle densities. Still, the increased computation burden causes significant latency when the scene complexity rises. In contrast, the proposed method and \cite{fan2025flying}, due to the fixed-shape sensing representation, their real-time performance is only slightly affected by the number of raw point clouds. The online optimization of \cite{lu2024fapp} enhances the safety by explicitly pushing the trajectories to the free space. But a large number of motion-variable obstacles elicit lagging state estimates, while frequent replanning is triggered but relies only on predictions discredited by compound errors. This results in unsafe behaviors in highly dynamic settings. The learning-based object-aware approach \cite{xu2025navrl} faces a similar dilemma. The sacrifice of agility realizes higher path efficiency and excellent static navigation performance. However, attributed to the strategy of learning avoidance maneuvers from dynamic-static compositions of the environment, its hyper-efficient and adaptive RL network policy is held back by its heavy detection-tracking modules and extra safety actions. This fundamentally leads to disadvantages in dynamic scenarios. 

\begin{figure}[t] \centering
	\includegraphics[width=0.485\textwidth]{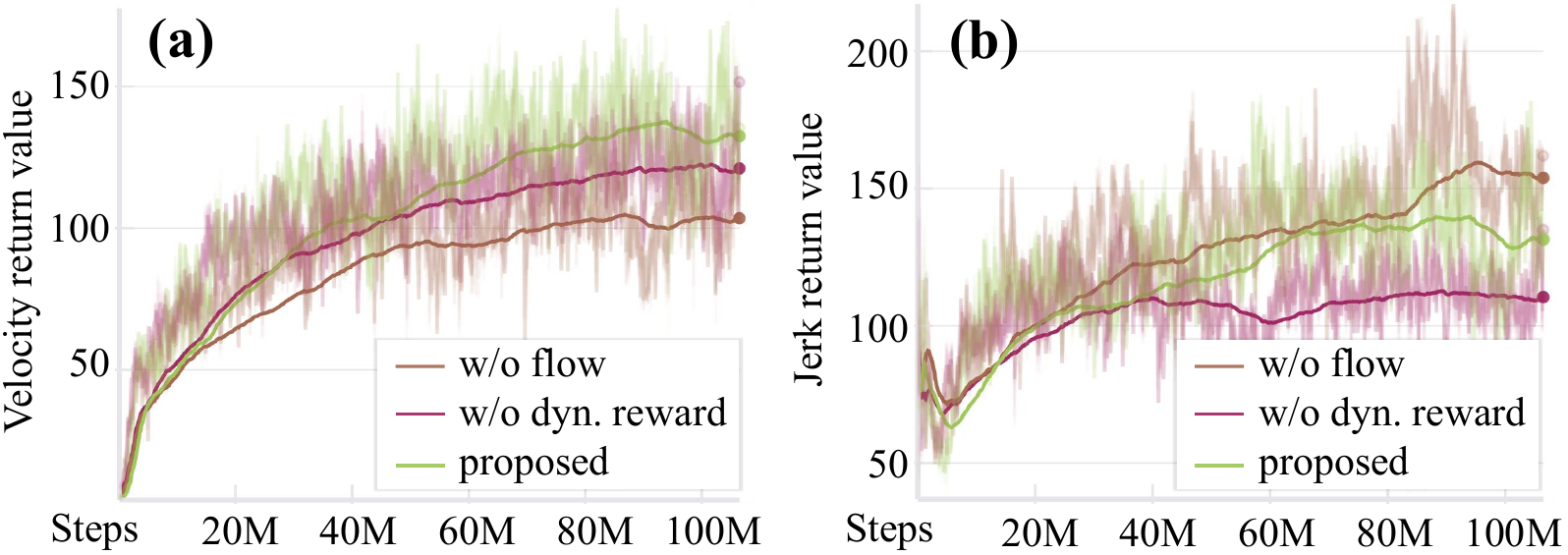}
	\caption{Reward components that show outstanding changes after ablation. The curves show rewards of the ablated systems after training the same steps as the proposed system, including the (a) velocity and (b) jerk rewards. They demonstrate the extent to which the agent is in the desired velocity interval and the extent to which the acceleration command mutates.}
	\label{fig_ablation_train}
\end{figure}
\begin{table}[t]
    \setlength{\tabcolsep}{5pt}
    \centering
    \caption{Ablation Study on Point Flow and Dynamic Obstacle Avoidance Reward with Different Obstacle Speeds}
    \begin{threeparttable}   
    {
    
    \begin{tabular}{ccccccc}
       \toprule
       \toprule
       \textbf{Obstacle Speed} & \textbf{Ablation} & \textbf{$\eta(\%)$} & \textbf{$\alpha_c$} & \textbf{$v_a(m/s)$} \\
       \hline
       
    \multirow{2.5}{*}{Low}
     & Ours & \best{90} & / & \secondbest{3.09} \\
    \multirow{2.5}{*}{($\| \boldsymbol{v}_{\text{dobs}} \| < 1\ m/s$)}
     & w/o flow& 60 & \secondbest{0.67} & 2.87 \\
     & w/o $r_{\text{dobs}}$ & \secondbest{70} & \best{0.78} & \best{3.73} \\
     \hline 
     
    \multirow{2.5}{*}{Medium}
     & Ours & \best{80} & / & \secondbest{3.08} \\
    \multirow{2.5}{*}{($\| \boldsymbol{v}_{\text{dobs}} \| < 2\ m/s$)}
     & w/o flow& 50 & \secondbest{0.63} & 2.68 \\
     & w/o $r_{\text{dobs}}$ & \secondbest{55} & \best{0.69} & \best{3.57} \\
     \hline 
     
    \multirow{2.5}{*}{High}
     & Ours & \best{60} & / & \secondbest{2.92} \\
    \multirow{2.5}{*}{($\| \boldsymbol{v}_{\text{dobs}} \| < 3\ m/s$)}
     & w/o flow& 35 & \secondbest{0.58} & 2.51 \\
     & w/o $r_{\text{dobs}}$ & \secondbest{40} & \best{0.67} & \best{3.38} \\
     \bottomrule
     \bottomrule 
    \end{tabular}
    }
    \end{threeparttable}
    \label{tbl:ablation_study}
\end{table}

Meanwhile, insights that discard intermediate modules show a high adaptability to moving objects. \cite{fan2025flying} learns directly from stacked historical scanning sequences, which can traverse dynamic clutters at high speeds. However, the lack of awareness of motion patterns between frames leads to difficulties in capturing threat differences across obstacle situations. This further gives rise to low safety and overfitting to dynamic scenarios, as well as breaking down in static scenarios. As can be seen from Table \ref{tbl:baseline_compare}, thanks to the change-aware LiDAR sensing representation and the flight skills empowered by relative motion-guided policy optimizaiton, our method exhibits the most reliable safety in the vast majority of scenes, maintaining a success rate of $40\%$ in the most complex environment containing both dynamic and static obstacles compared to the less than $30\%$ of the other methods. Furthermore, in comparison with \cite{fan2025flying}, although the agility of our method is not as high, it has the advantage of being adaptively regulated by the environment complexity based on the reference input.

\begin{figure*}[t] \centering
	\includegraphics[width=1.0\textwidth]{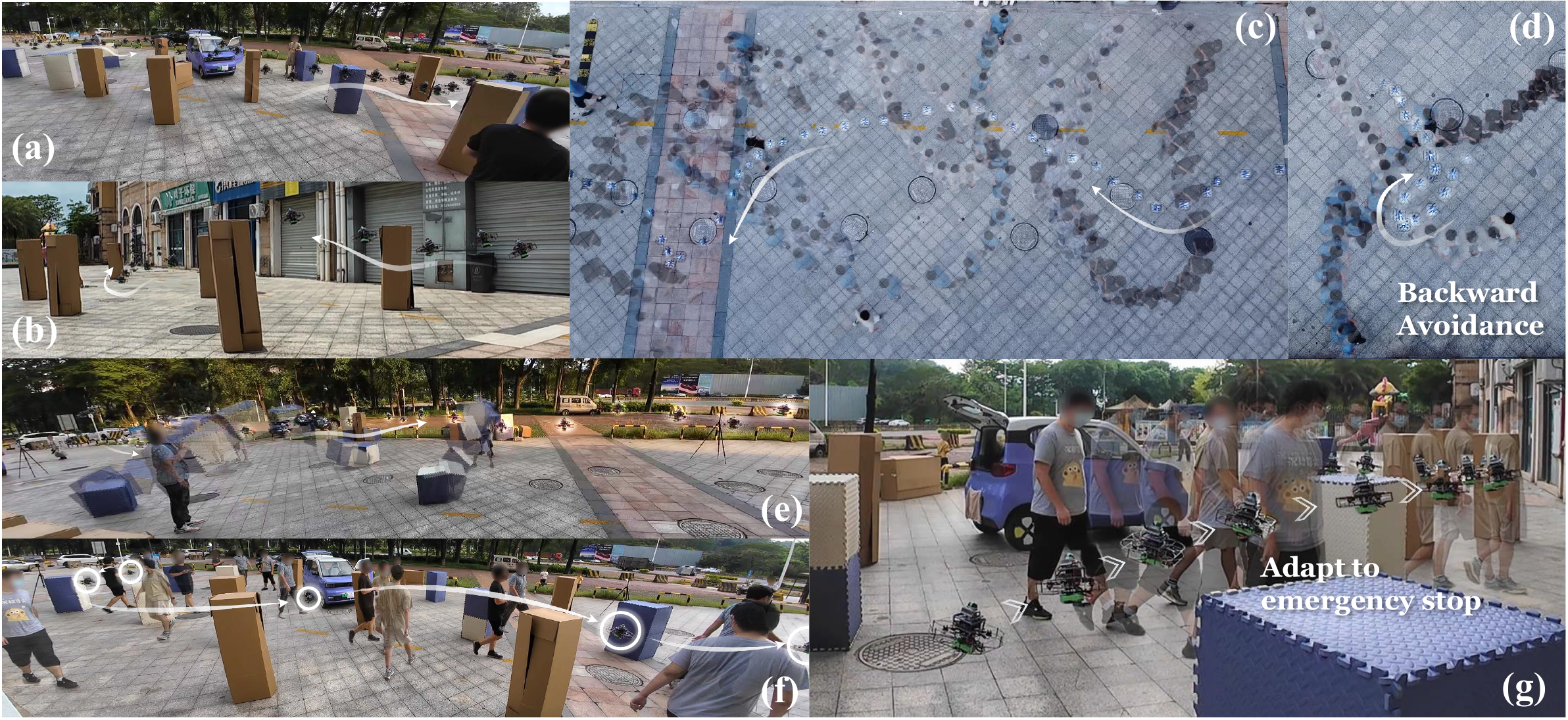}
	\caption{Real-world experiments. (a) Traversal through a box environment. (b) Agile flight with high reference acceleration input. (c) Frequently interacting with pedestrians. (d) Highlight of a backward avoidance when performing a similar experiment to (c) in a lower agility. (e) Reacting to fast-moving objects. (f) Navigation among static and dynamic obstacles. (g) Details of adapting to an obstacle emergency stop in experiment (f).}
	\label{fig_real_world}
\end{figure*}

\emph{2) Ablation on method components}:
We perform ablations on the point flow and the dynamic obstacle avoidance reward $r_{\text{dobs}}$, and the ablated training results are shown in Fig. \ref{fig_ablation_train}. We conduct experiments in dense scenarios containing $33$ dynamic obstacles and implement control groups with different obstacle speed limits, which helps to directly assess the ablation impact on dynamic obstacle avoidance performance. The results are shown in Table \ref{tbl:ablation_study}, where the newly introduced metric $\alpha_c$ represents the ratio of the success rate of the ablated system to the success rate of the proposed system.

Leveraging the intuitive depth sensing, the system with ablated point flow still ensures the basic safety, but the loss of relative motion perception means it can only keep reacting to the dynamic obstacles under the static assumptions. The difficulty in taking efficient and early avoidance actions directly leads to the decrease of flight speed and the reduction of velocity reward during training. The system without dynamic obstacle avoidance reward can not effectively correlate the estimated environmental changes with the desired behavior. It struggles to realize the differences in threat prioritization when handling a multi-dynamic obstacle scenario, thus it prefers to give over-excited maneuvers with sudden changes to achieve avoidance when the safety distance is very small. This causes a drop in jerk reward and also results in an inappropriately excessive agility. In a general view, the effects of the ablated system components on flight safety become more obvious as the dynamics of the environment increase, which are manifested by the gradual decrease of $\alpha_c$, which reflects the proximity in reliability. 

\subsection{Real-World RL Deployment}
Thanks to the kinematic-level observation and action designs, we can use a physical quadrotor that has different dynamics parameters from training. We apply the policy obtained from the sensor attitude-ignored training to the rigidly connected LiDAR-quadrotor in reality, where the Livox Mid-360 LiDAR is used to provide point cloud perception and the PX4 flight controller is used to track the desired acceleration as in Fig. \ref{fig_system_overview}. We use TensorRT for inference acceleration on the on-board computing unit Nvidia Jetson Orin NX, which ensures a control frequency of $50\ \text{Hz}$.

\subsection{Real-World Experiments}
\emph{1) Traversal through a box environment \& Agile flight with high reference acceleration input}: To evaluate the obstacle avoidance ability in static environments, we design a scene composed of boxes placed horizontally or vertically at different heights, which is similar to the column obstacle in training. As shown in Fig. \ref{fig_real_world}(a), the quadrotor can make adaptive decisions to fly through the free space on either side or above the obstacle. In order to test the ability of agile flight, we retain a few columnar obstacles and increase the expected agility to a high level. As shown in Fig. \ref{fig_real_world}(b), the quadrotor can nimbly shuttle among obstacles and reach a maximum speed of $14.57\ \text{km/h}$  ($4.05\ \text{m/s}$). 

\emph{2) Frequently interacting with pedestrians \& Reacting to fast-moving objects}: To assess the ability of handling dynamic objects, we arrange for multiple pedestrians to walk in chaotic and subjectively varying trajectories and speeds, which is different from the uniform linear motion of dynamic obstacles in training. As shown in Fig. \ref{fig_real_world}(c), the quadrotor can efficiently fly through areas without potential dangers. When the threat of the dynamic environment is significant, the policy does not greedily insist on the fastest-forward maneuvers. Instead, it chooses actions such as deceleration or stopping to ensure safety. In a similar environment but lower expected agility setup, backward action can also be triggered to achieve a low-speed avoidance solution, as shown in Fig. \ref{fig_real_world}(d). In order to test the response to fast-moving obstacles, we throw boxes along the path ahead of the quadrotor to interfere with its normal flight, as shown in Fig. \ref{fig_real_world}(e), which is similar to the scenario of a fast-moving obstacle appearing from the side in training. The quadrotor can continuously respond to multiple fast-moving thrown obstacles and always maintain a safe distance from them. 

\emph{3) Navigation among static and dynamic obstacles \& Flying in natural clutters with pedestrians}: To evaluate the navigation ability in a complex dynamic environment, we design a scene containing columns and pedestrians. The dense occlusion and unpredictable dynamic behaviors pose challenges for perception and planning. As shown in Fig. \ref{fig_real_world}(f), the quadrotor repeatedly adjusts its maneuverability and travel direction to safely bypass the dynamic and static obstacles. Meanwhile, when the speed of the dynamic obstacle undergoes a sudden change, such as an emergency stop in Fig. \ref{fig_real_world}(g), the policy can still handle it robustly without being triggered to take excessive actions. We also deploy the system among unevenly distributed trees while integrating pedestrians into the environment. The challenges posed by natural clutters primarily stem from the domain gap between training, including highly unstructured observations, a significant amount of uninformative points, and unaccounted-for wind disturbances. As shown in Fig. \ref{fig_cover}, in this long-distance task, the policy can be easily generalized to tree scenes and navigate flexibly through pedestrians and trees. 

\subsection{Discussions on Robustness and Limitations}
The robustness of the system is reflected in the performance under sensor degradation and unseen motion patterns. (i) The point flow yields reliable estimates under noise and sensor occlusions, whereas the degraded distance map may affect policy performance. (ii) Owing to the immediate and relative perception nature of the point flow, the policy can achieve robust near-distribution generalization to unseen obstacle motions with novel trajectories and speeds.

The main limitation of the system is reflected in some failure cases. (i) Due to the limitation of the sensor FOV and training obstacle setting, avoiding obstacles from above is not always successful. (ii) Due to the loss of dynamic perception caused by occlusion, collisions may happen with the high-speed obstacle that emerges beside others. (iii) Due to insufficient training sample coverage, collisions may happen with the obstacles with abrupt relative motion trend changes. In future work, we will incorporate risk awareness to overcome out-of-view and occluded threats, and supplement urgent maneuvers learning by diversifying training conditions.
\section{Conclusion}
In this work, we showed that combining single LiDAR sensing with RL can realize an end-to-end autonomous flight system with effortless sim-to-real transfer. We demonstrated that, without introducing object detection, tracking, and prediction modules, lightweight access to complex dynamic environments can be achieved through the encoding of point cloud data and the extraction of multi-frame motion features. And we exhibited that, without trajectory generation, the dynamic obstacle avoidance behaviors can be implicitly driven by the relative motion-guided policy optimization. 


%

\ifCLASSOPTIONcaptionsoff
  \newpage
\fi
\bibliographystyle{IEEEtran}
\bibliography{ref}











\end{document}